\documentclass[runningheads]{llncs}
\usepackage{graphicx}
\usepackage{booktabs}       
\usepackage{amsfonts}       
\usepackage{hyperref}       
\usepackage{caption}
\usepackage{subcaption}
\usepackage{amsmath}
\usepackage{graphics} 
\begin{document}
\title{Hotel Recognition via Latent Image Embeddings}
%
%
\author{Boris Tseytlin \inst{1}\orcidID{0000-0001-8553-4260} \and Ilya Makarov \inst{1}\orcidID{0000-0002-3308-8825}}
\authorrunning{B. Tseytlin, I. Makarov.}

\institute{HSE University, Moscow, Russia \\
\email{b.tseytlin@lambda-it.ru,iamakarov@hse.ru}
}

\maketitle              
\begin{abstract}
We approach the problem of hotel recognition with deep metric learning. We overview the existing approaches and propose a modification to Contrastive loss called Contrastive-Triplet loss. We construct a robust pipeline for benchmarking metric learning models and perform experiments on Hotels-50K and CUB200 datasets. Contrastive-Triplet loss is shown to achieve better retrieval on Hotels-50k.

\end{abstract}
\section{Introduction}

In large-scale image-based hotel recognition, the goal is to retrieve hotels from a large database of hotel images. The input is a hotel photo and the retrieved hotel images are expected to be from the same hotel or hotel chain. We approach this task as a deep metric learning problem. In this setting, we train a deep neural network to project images into an informative low-dimension latent space. This space is used for approximate nearest-neighbor retrieval. 

The paper is structured in the following way. First we carefully investigate the available approaches. We investigate failure cases of Contrastive and Triplet losses, where they miss out on potentially important information within a triplet. Based on the hypotheses that Triplet loss and Contrastive loss use different information available in a batch we propose the Contrastive-Triplet loss. We construct a benchmarking pipeline for robust model comparisons, ensuring equal conditions for all approaches. The Mean Average Precision at R (MAP@R) metric is used for evaluation. Finally, we experimentally test the proposed approach by comparing the performance of Contrastive-Triplet loss to individual Contrastive and Triplet losses on the CUB200 and Hotels-50k datasets. We finish with a discussion of experimental results. The code for reproducing our experiments is available on Github.\footnote{https://github.com/btseytlin/metric\_benchmarks}

Our contributions are two-fold:
\begin{enumerate}
    \item We find examples where Contrastive and Triplet losses fail to include important information within a batch and formulate the hypothesis that the classic losses use different aspects of similarity. 
    \item We propose the Contrastive-Triplet loss function which combines Contrastive loss and Triplet loss with no additional computational overhead.
    \item For the first time to our knowledge we evaluate a newly proposed Deep Metric Learning approach against the fair benchmarking pipeline by \textit{Musgrave et al.} \cite{musgrave2020metric} and open-source the code for reproduction.
\end{enumerate}

\section{Related work}
\label{sec:related_work}

\begin{figure}
\centering
\begin{subfigure}{\textwidth}
      \centering
      \includegraphics[width=\linewidth]{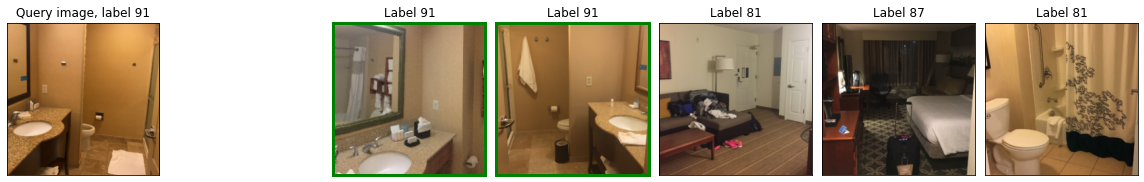}
      \caption{Successful retrieval example.}
      \label{fig:retrieval_example_good}
\end{subfigure}%
\hfill
\begin{subfigure}{\textwidth}
  \centering
  \includegraphics[width=\linewidth]{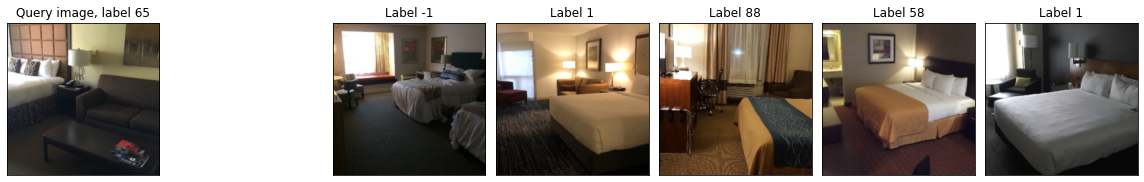}
  \caption{Unsuccessful retrieval example.}
  \label{fig:retrieval_example_bad}
\end{subfigure}
\caption{Retrieval examples for Hotels-50k chain recognition using Contrastive-Triplet loss.}
\label{fig:retrieval_examples}
\end{figure}

The tasks of image retrieval, scene recognition and place recognition can be solved using metric learning. Metric learning attempts to map data to an embedding space, where similar data are close together and dissimilar data are far apart. There are two main approaches to deep metric learning: embedding and classification losses. 

\subsection{Classification} In this setting, a classification model is trained using softmax loss and the embeddings produced by the penultimate layer of the model are used.

The first caveat of this approach is that a model trained for classification does not directly optimize for similarity, which sometimes leads to sub-par results. The second caveat is that image retrieval problems tend to have large numbers of semantic labels, which makes it computationally hard to train a classifier model. Despite that, \textit{Zhai et al.} \cite{zhai2018classification} modify a softmax loss trained classifier for metric learning and show that such approach is very competitive and can be scaled to large numbers of classes.

\textit{Deng et al.} \cite{deng2019arcface} address the problem that softmax loss does not directly optimize for similarity. They introduce a new loss function ArcFace loss that distributes the data points on a hypersphere. Unlike a generic softmax loss, this loss function directly enforces an angular margin $m$ between embeddings of different classes. Authors achieve great results on face recognition benchmark datasets, which tend to have high inter-class variance and large numbers of clusters. 

\subsection{Embedding losses} 
In this approach a model is trained to explicitly learn an embedding of data into a latent space, where the similarity is maximized for semantically similar data points and minimized otherwise. This is usually achieved by considering pairs, triplets or N-tuplets of images.

Contrastive loss \cite{hadsell2006dimensionality} is a classic approach that had wide adoption. The idea behind it is to analyze pairs of embeddings and penalize the embedder model for images of the same label being too far apart and for images of different labels being too close. While training with Contrastive loss, the loss is computed over pairs of images within a batch. A pair with the same label is called a positive pair. A pair with different labels is called a negative pair. The loss is calculated as follows:
\begin{equation}  \label{Contrastive}
L_{Contrastive} = [d_p - m_{pos}]_+ + [m_{neg} - d_n]_+
\end{equation}
Where $d_p$ is the distance for the positive pair, $d_n$ is the distance for the negative pair, $m_{pos}$ and $m_{neg}$ are hyperparameters, $[x]_+ = max(0, x)$ is the hinge function.

Another classic approach is Triplet loss \cite{weinberger2009distance}. Triplets of images $(a, p, n)$ are selected from each batch, where $x_a$ is an anchor image, $p$ has the same label as the anchor, $n$ has a different label. Triplet loss minimizes the anchor-positive pair distance and maximizes the anchor-negative distance, so that the anchor-positive pair is closer than the anchor-negative pair by some margin $m_{triplet}$:
\begin{equation}  \label{Triplet}
L_{Triplet} = [d(a, p) - d(a, n) + m_{triplet}]_+
\end{equation}

Both Triplet loss and Contrastive loss share the issue of selecting useful triplets or pairs within a batch. The total number of possible triplets in a dataset is $O(n^3)$, but only a tiny fraction of these produce useful gradients. Triplet mining aims to select useful triplets for learning. 

\textit{Xuan et al.} \cite{xuan2020improved} show that sampling the easiest positive and semi-hard negative pairs for triplets leads to better results on datasets with high intra-class variance. An easiest positive image is the closest positive to the anchor in the batch. A semi-hard negative is a negative image that is further from the anchor than the positive image, but within the triplet margin. Such selection pushes each image closer to the most similar image of the same class. This reduces the over-clustering problem, when very different images of the same class are mapped to the same place. Authors indicate that mapping each image close to the closest positive leads to better generalization on unseen data.

\textit{Movshovitz et al.} \cite{movshovitz2017no} argue that it's not enough to select useful triplets from each training batch, but careful sampling of batches is a problem as well. To counter the sensitivity to individual triplets picked into the batch, they propose to learn a small set of proxy points $P$ that approximates the original data distribution. In a supervised setting, where all data points are labelled as belonging to a class, a single proxy is used for each class. Authors extend the NCA loss \cite{goldberger2005neighbourhood} with proxies, proposing the Proxy-NCA loss. This has the additional advantage over Triplet loss as the margin hyperparameter is no longer needed. This approach was extended \cite{wern2020proxynca++} to ProxyNCA++ later.


\textit{Wang et al.} \cite{wang2019multi} claim that different ranking losses take in account different aspects of similarity. Contrastive loss is based on the similarity between the anchor and the negative example. Triplet loss is based on the difference between the anchor-negative similarity and similarity to positive examples. According to the author each method only uses a specific part of information about a tuple. They propose a multi-similarity loss that attempts to capture all of the information at once and claim that it naturally selects harder pairs.

\subsection{Other approaches} 

\textit{Kim et al.} \cite{kim2018attention} propose an attention-based ensemble model. Multiple attention masks are used after spatial feature extraction, which focuses the model on different regions of the input image. They utilize a divergence loss that promotes each attention mask to learn a different embedding. Similar methods are used for content-based retrieval \cite{tseytlin2020content} and automated content editing techniques \cite{golyadkin2020semi,lomotin2020automated}.

There were attempts to combine embedding and classification losses for better performance \cite{lomov2021fault}. \textit{Jun et al.}\cite{jun2019combination} combine the two major approaches by pre-training a DCNN using a classification loss and then fine-tuning the network using Triplet loss. 

An loss-agnostic way to improve model performance is to provide the model more useful training samples. \textit{Lin et al.} \cite{lin2018deep} use a generator network to generate hard negative points from easy negatives, so that the whole distribution of negative samples is covered by the network during training. \textit{Ko et al.} \cite{ko2020embedding} take a different approach for generating hard points: they generate points by linearly interpolating between negative data points during training and picking the hardest point. Other approaches may directly learn metric space transformation via graph embeddings \cite{makarov2021survey,makarov2021fusion}.


\subsection{Reality check} 

\textit{Musgrave et al.} \cite{musgrave2020metric} benchmark the models proposed by numerous metric learning papers and find out that much of the claimed progress in the field was caused by flawed experiments. They find that most methods perform roughly like the classic Contrastive loss, Triplet loss or softmax loss approaches. Most of the gains of recent approaches are explained by using newer backbone DNN architectures, optimizing hyper-parameters on test subsets and using flawed accuracy metrics.

\section{Proposed improvements}

After analysing previous works on this topic, we formulated the following hypothesis that could be experimentally tested: embedding losses use different aspects of similarity and current embedding losses do not use all of the available information within a batch.

\subsection{Contrastive-Triplet loss}

To test the hypotheses that different embedding losses use different information in a batch, we engineer some examples where the classic losses could fail.

\begin{figure}
\centering
\begin{subfigure}{.5\textwidth}
  \centering
  \includegraphics[width=.7\linewidth]{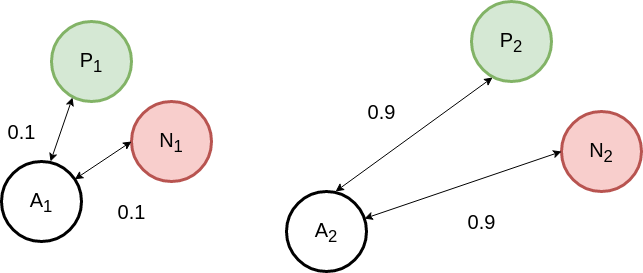}
  \caption{Failure case for Triplet loss}
  \label{fig:failure_cases:a}
\end{subfigure}%
\begin{subfigure}{.5\textwidth}
  \centering
  \includegraphics[width=.7\linewidth]{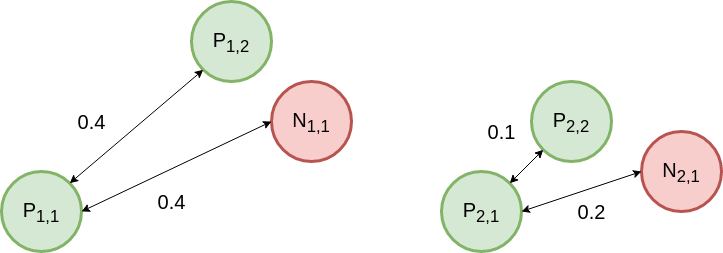}
  \caption{Failure case for Contrastive loss}
  \label{fig:failure_cases:b}
\end{subfigure}
\caption{Different triplets for which Triplet loss (a) or Contrastive loss (b) produce the same loss value, potentially ignoring important properties of triplets.}
\label{fig:failure_cases}
\end{figure}

\textit{Example where Triplet loss fails}. Consider Triplet loss with $m_{triplet} = 0.1$ and two triplets with the following distances: $d(a_1, p_1) = 0.9$, $d(a_1, n_1) = 0.9$, and $d(a_2, p_2) = 0.1$, $d(a_2, n_2) = 0.1$. For both triplets the loss is equal to $0.1$, even though they are different. The information about the positive example being very far from the anchor is ignored in the first triplet. This case is illustrated by Figure \ref{fig:failure_cases:a}.

Another failure case of Triplet loss is when a triplet produces no gradients. For example consider a triplet with $d(a, p) = 0.3$, $d(a, n) = 0.4$. This is considered an "easy" triplet and produces zero loss, even though the positive example could be pulled closer and the negative could be pushed further away. 

\textit{Example where Contrastive loss fails}. Consider Contrastive loss $m_{pos} = 0.2$, $m_{neg} = 0.5$ and pairs with the following distances: $d(p_{1,1}, p_{1,2}) = 0.4$, $d(p_{1,1}, n_{1,1}) = 0.4$, and $d(p_{2,1}, p_{2,2}) = 0.1$, $d(p_{2,1}, n_{2,1}) = 0.2$. For both cases the resulting loss is $0.3$, even though in the first case the distances are equal and the example is therefore harder. The information about the relative distance between the positive and the negative examples is ignored. This is illustrated by Figure \ref{fig:failure_cases:b}.

We attempt to create a loss function for which these failure cases are avoided. We propose a new loss function that combines Contrastive and Triplet loss to use more information about the embedding whilst providing no computational overhead.

Consider a triplet $(a, p, n)$. In addition to Triplet loss, we can also calculate the Contrastive loss for the triplet by taking $d_p = d(a, p)$ and $d_n = d(a, n)$.

By summing these losses, we obtain the Contrastive-Triplet loss:
\begin{multline}  \label{ContrastiveTriplet}
L_{ContrastiveTriplet} = [d(a, p) - m_{pos}]_+ [m_{neg} - d(a, n)]_+ \\ + \alpha_{triplet} [d(a, p) - d(a, n) + m_{triplet}]_+
\end{multline}

Where $\alpha_{triplet}$ is a hyperparameter weight for the Triplet loss component.

The pairwise distances matrix can be reused when computing both loss components, so there is no computation overhead.

\section{Experiments}
\label{sec:experiments}

\begin{table}[t]
\centering
\caption{Optimal hyperparameters of approaches for the CUB200 dataset (top) and the Hotels-50k dataset chain recognition task (bottom) found via Bayesian optimization. Dashes mark hyperparameters that are not used by an approach.}
\label{table:cub_hyperparams}
\resizebox{\columnwidth}{!}{%
    \begin{tabular}{ |c|c|c|c|c|c|c|c|c|c| } 
     \hline
     Approach & $m_{pos}$ &  $m_{neg}$ & $m_{triplet}$ & $\alpha_{triplet}$ & $m_{own}$ & $m_{other}$ & $\alpha_{own}$ & $\alpha_{other}$ & $delay$ \\ 
     \hline
     Batch-hard Triplet loss (baseline) & - & - & 0.289 & - & - & - & - & - & - \\ 
     Contrastive loss & 0.330 & 0.691 & - & - & - & - & - & - & - \\ 
     Contrastive-Triplet & 0.030 & 0.742 & 0 & 1 & - & - & - & - & - \\ 
     \hline
    \end{tabular}
}

\medskip

\label{table:hotels_hyperparams}
\resizebox{\columnwidth}{!}{%
    \begin{tabular}{ |c|c|c|c|c|c|c|c|c|c| } 
     \hline
     Approach & $m_{pos}$ &  $m_{neg}$ & $m_{triplet}$ & $\alpha_{triplet}$ & $m_{own}$ & $m_{other}$ & $\alpha_{own}$ & $\alpha_{other}$ & $delay$ \\ 
     \hline
     Batch-hard Triplet loss (baseline) & - & - & 0.396 & - & - & - & - & - & - \\ 
     Contrastive loss & 0.111 & 0.407 & - & - & - & - & - & - & - \\ 
     Contrastive-Triplet & 0.080 & 0.989 & 0.608  & 0.884 & - & - & - & - & - \\ 
     \hline
    \end{tabular}
}
\end{table}

We follow the fair comparison procedure proposed by \textit{Musgrave et al.} \cite{musgrave2020metric} and use the proposed benchmarker software \cite{Musgrave2019} for our experiments. We ensure that all approaches are tested using the same backbone network, BatchNorm parameters, image augmentation and other conditions. The hyperparameters for each approach are optimized via Bayesian optimization first. We also use cross-validation and multiple reproductions to obtain robust estimates of model performance as well as confidence intervals. We use 25 Bayesian optimization iterations for CUB200 and 10 for Hotels-50k instead of 50 due to computation resource limits and time constraints.

\subsection{Experiment setup}
Experiments use the PyTorch \cite{NEURIPS2019_9015} library with the following procedure:
\begin{itemize}
  \item The trunk model is an ImageNet \cite{deng2009imagenet} pre-trained BN-Inception \cite{ioffe2015batch} network.
  \item The head model is a 1-layer MLP. 
  \item Trunk BatchNorm parameters are frozen during training.
  \item Embedding dimensionality is 128.
  \item Batch size is 32. Batches are constructed by sampling 8 classes and randomly sampling 4 images for each class.
  \item During training, images are first resized so that the shorter side has length 256, then a random crop is made of size between 40 and 256, and aspect ratio between 3/4 and 4/3. Finally, the crop is resized to 227x227, and flipped horizontally with 50\% probability. 
  \item During evaluation, images are resized to 256 and then center cropped to 227.
  \item We use separate RMSprop optimizers with a learning rate of $1e-6$ for trunk and head networks. 
  \item Embeddings are L2 normalized.
  \item Losses are reduced by averaging all non-zero loss values.
  \item In case of CUB200, the first half of classes are used as the training set. The second half of classes are used as the test set. In case of Hotels-50k the train-test split is predefined.
  \item We use 4-fold cross-validation on the training set. Each fold consists of $1/4$ training classes and all folds are class-disjoint.
  \item Training stops when validation accuracy does not improve for 9 consecutive epochs. 
  \item We perform 5 reproduction training runs with the best hyperparameters. Within each reproduction, for each CV split, we load the best model based on validation set error. We use the best models to obtain train and test set embeddings. We concatenate these embeddings into 512-dim embeddings and L2-normalize them. Using these embeddings we test retrieval and report the mean accuracy and confidence intervals.
\end{itemize}

For Hotels-50k the procedure has the following differences:
\begin{itemize}
    \item 10 iterations of Bayesian optimization are performed.
    \item Only 500 batches are processed per epoch.
    \item For each cross-validation fold, the validation fold only has labels that are present in the training set. This is done to mimic the task of the dataset.
\end{itemize}

Batch-size can have a significant effect on neural networks, but \textit{Musgrave et al.} \cite{musgrave2020metric} demonstrate that increasing  batch size from 32 to 256 has little effect to relative rankings of metric learning approaches.

\subsection{Accuracy metrics}
We compare solutions using Mean Average Precision at R (MAP@R).
For a given query it's defined as follows:
\begin{equation}  \label{mapr}
MAP@R = \frac{1}{R} \sum^{R}_{i=1} P(i)
\end{equation}

Where $P(i) = \text{precision at i} \ \text{if the i-th retrieval is correct and} \ 0 \ \text{otherwise}$. $R$ is the total number of references that have the same label as the query. 

This metric takes in account both the number of correct retrievals and their ranking. We also provide the commonly used $P(1)$ metric, also called Top-1 Accuracy, for comparison. However it should be noted that the metric can be very high even for poorly-clustered samples, as demonstrated in \cite{musgrave2020metric}.

\subsection{Evaluation}

\begin{figure}
\centering
\begin{subfigure}{.45\textwidth}
  \centering
  \includegraphics[width=\linewidth]{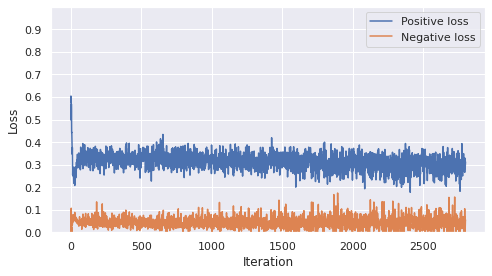}
  \caption{Contrastive loss history}
  \label{fig:contrastive_vs_contrastive_triplet_losses:sub1}
\end{subfigure}%
\begin{subfigure}{.45\textwidth}
  \centering
  \includegraphics[width=\linewidth]{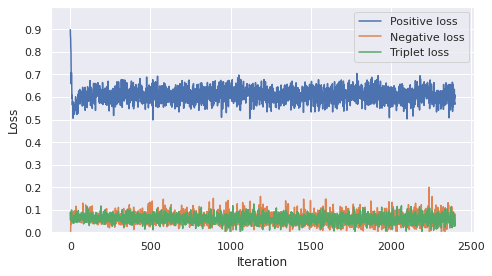}
  \caption{Contrastive-Triplet loss history}
  \label{fig:contrastive_vs_contrastive_triplet_losses:sub2}
\end{subfigure}
\caption{Loss histories for Contrastive loss (a) and Contrastive-Triplet loss (b) when training on CUB200 with the best hyperparameters. The negative loss value is approximately same for both approaches. Positive loss changes considerably in presence of the Triplet loss component. The overall convergence pattern remains the same. }
\label{fig:contrastive_vs_contrastive_triplet_losses}
\end{figure}

For the best hyperparameters, 5 reproduction iterations are run, following the standard cross-validation scheme. For each reproduction the accuracy score is computed. For each CV partition within a reproduction, the highest-accuracy checkpoint is loaded. The 4 models are used to obtain embeddings. For both train and test sets we concatenate the 128-dim embeddings of the 4 models to get 512-dim embeddings, and then L2 normalize them. We test and report the accuracy of retrieval using these embeddings. In the end, we aggregate the scores of reproductions to obtain mean accuracy metrics and confidence intervals.

\subsection{CUB200 dataset}

We benchmark our approaches on CUB200-2011 \cite{WelinderEtal2010} dataset. The dataset contains photos of birds taken in the wild, with bird species used as labels. It's a commonly used benchmark in metric learning. Just like Hotels-50k, it contains a lot of hard cases of two types:
\begin{itemize}
    \item images of the same class that are very different visually,
    \item images of different classes that are very similar visually.
\end{itemize}
In other words, this dataset is notable for high intra-class variance and high similarity between some classes.  

The dataset contains 11,788 images distributed between 200 classes. In accordance with previous literature, we use the first 100 classes for the training dataset, and the last 100 classes for the test dataset. 

\subsection{Hotels-50k dataset}

Finally, we benchmark on the Hotels-50k \cite{hotels50k} dataset. The dataset contains images taken in hotel rooms. Some images are non-professional photos of hotel rooms taken with the TraffickCam app, the rest are professional photos from hotel booking websites. Each photo is labelled with hotel id, hotel chain id and image source (TraffickCam or not). All test images come from TraffickCam, so the task is to identify a hotel or chain given a poor-quality user submitted photo. The dataset provides two tasks: hotel instance recognition and hotel chain recognition. We only approach the hotel chain recognition task.

Due to the large size of the dataset, time and resource constraints, we use a subset of the dataset. For the training set, we remove all hotel chains that have fewer than 100 images and all hotels that have fewer than 10 images. Among the remaining images, we keep only a sample the non-TraffickCam images, such that their number is equal to the number of TraffickCam images. Finally, we remove all hotels from the test set if they are not in the resulting training set.

In the end, the training subset contains 86,936 images from 30,467 hotels of 90 hotel chains. 43,468 images among these come from TraffickCam. The test subset contains 10,900 images from 3,244 hotels of 85 hotel chains. All images in the testing set come from TraffickCam. The final subset we used, as well as the code for recreating it, can is available on Github \footnote{https://github.com/btseytlin/metric\_benchmarks}.

To speed up training on this large dataset, we convert the test and train sets into Lightning Memory Mapped Database \cite{lmdb} binary files storing the following key value pairs: $(image\_index, (image, chain\_id, hotel\_id, image\_source))$. LMDB allows multiple concurrent readers to read images from the file. This significantly increases the speed of loading images and training.

\subsection{Baseline}

For comparison, we reproduce the approach used in the original Hotels 50K paper \cite{hotels50k}. A neural network is trained using Triplet loss. Batch-hard mining is used: a miner that samples the hardest positive and hardest negative triplet for each anchor. All losses are averaged, including zero values. All other parameters, including backbone network, batch size, embedding dimension and image augmentations, are the same as for other approaches (described in Section \ref{sec:experiments}), to ensure a fair comparison.

\section{Results}

The benchmark results can be found in Table \ref{table:cub_results} and Table \ref{table:hotels_results_chains}. The optimal hyperparameters for the CUB200 dataset can be found in Table \ref{table:cub_hyperparams}, and for the Hotels-50k dataset in Table \ref{table:hotels_hyperparams}.

The hypotheses behind Contrastive-Triplet loss was that the different components of the loss might describe different aspects of similarity. If that was true, combining them with optimized hyperparameters would achieved better results. 

\begin{table}[!ht]
\centering
\caption{Benchmark results on CUB200.}
\label{table:cub_results}
\resizebox{\columnwidth}{!}{%
    \begin{tabular}{ |c|c|c|c|c| } 
     \hline
     Approach & MAP@R & MAP@R confidence & P(1) & P(1) confidence \\ 
     \hline
     Untrained backbone & 0.134 & - & 0.524 & - \\ 
     Triplet loss (baseline) & 0.243 & [0.237, 0.249] & 0.647 & [0.639, 0.655] \\
     Contrastive & \textbf{0.260} & [0.257, 0.264] & 0.668 & [0.664, 0.671] \\ 
     Contrastive-Triplet & 0.258 & [0.254, 0.262] & 0.670 & [0.665, 0.675] \\
     \hline
    \end{tabular}
}

\caption{Benchmark results on Hotels-50k chain recognition.}
\label{table:hotels_results_chains}
\resizebox{\columnwidth}{!}{%
    \begin{tabular}{ |c|c|c|c|c| } 
     \hline
     Approach & MAP@R & MAP@R confidence & P(1) & P(1) confidence \\ 
     \hline
     Untrained backbone & 0.023 & - & 0.175 & - \\ 
     Triplet loss (baseline) & 0.029 & [0.028, 0.030] & 0.255 & [0.251, 0.260] \\
     Contrastive & 0.031 & [0.030, 0.033] & 0.268 & [0.262, 0.275] \\ 
     Contrastive-Triplet & \textbf{0.039} & [0.037, 0.040] & 0.300 & [0.290, 0.310] \\
     \hline
    \end{tabular}
}
\end{table}

\begin{figure}[!ht]
\centering
\begin{subfigure}{.45\textwidth}
  \centering
  \includegraphics[width=\linewidth]{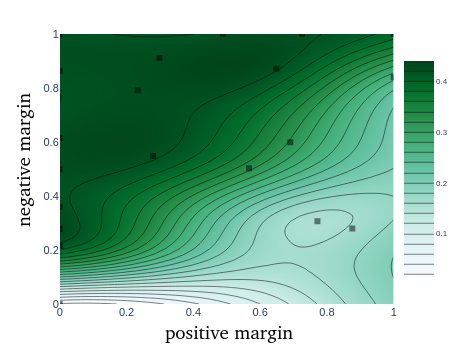}
  \caption{Contrastive loss Bayesian optimization plot}
  \label{fig:contrastive_vs_contrastive_triplet_bayesian:sub1}
\end{subfigure}%
\begin{subfigure}{.45\textwidth}
  \centering
  \includegraphics[width=\linewidth]{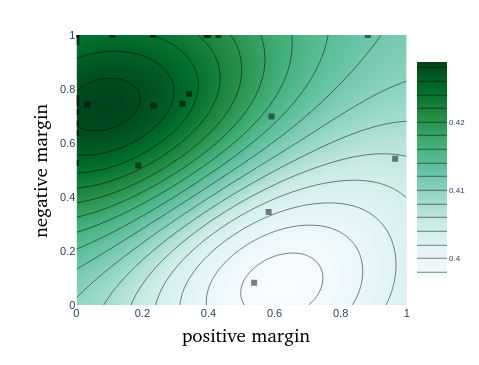}
  \caption{Contrastive-Triplet loss Bayesian optimization plot}
  \label{fig:contrastive_vs_contrastive_triplet_bayesian:sub2}
\end{subfigure}
\caption{Bayesian optimization plots for Contrastive and Contrastive-Triplet losses on dataset CUB200. Black squares indicate Bayesian trials. Darker colors indicate higher validation set MAP@R. Optimal values for positive and negative margin are much more stable in presence of Triplet loss component.}
\label{fig:contrastive_vs_contrastive_triplet_bayesian}
\end{figure}

It does not appear to be the case for CUB200. Loss history plots, shown on Figure \ref{fig:contrastive_vs_contrastive_triplet_losses}, indicate that adding a Triplet loss component does not significantly affect convergence. The negative loss component does not change with addition of the Triplet loss component. Our experiments show that adding a Triplet loss component to Contrastive loss shifts the positive loss margin towards zero, making the positive loss greater in value, while the Triplet loss margin also shifts towards zero. When $m_{triplet}$ is close to zero, Triplet loss only provides penalties when the positive pair is farther away than the negative pair. In that case Triplet loss component pulls positive examples that are too far away closer to the anchor, almost like positive Contrastive loss. For the CUB200 dataset, a combination of Triplet loss and Contrastive loss achieves the same results as Contrastive loss alone. This leads us to believe that either Triplet loss or Contrastive positive loss are redundant in this case. It should be noted that while Contrastive-Triplet loss does not achieve better results on CUB200, the hyperparameter values are much more stable for it than for Contrastive loss, as shown on Figure \ref{fig:contrastive_vs_contrastive_triplet_bayesian}.

However, the approach does achieve slightly better results on Hotels-50k dataset. This might be explained by the fact that Contrastive-Triplet loss has more stable hyperparameters. Fewer Bayesian optimization iterations were done for Hotels-50k, which could lead to better results for the method that takes less steps to optimize hyperparameters. Examples of retrieval are provided on Figure \ref{fig:retrieval_examples}. Overall, qualitative analysis indicates that the embedding produced is densely packed, with the distances between random pairs of images being very small, which leads to a lot of false positives. The network is not able to enforce a large margin between hotel chains.

\section{Conclusion}

In this paper we approached the task of hotel recognition. Having reviewed the existing approaches we investigated the failure cases of Contrastive and Triplet losses and proposed the Contrastive-Triplet loss. We constructed a robust experiment pipeline to ensure consistent and reproducible results. Our experiments indicate that Contrastive-Triplet loss doesn't outperform the baseline on CUB200, but achieves better retrieval results on Hotels-50k than baseline Contrastive and Triplet losses. 

\bibliographystyle{splncs04}  
\bibliography{references} 
\end{document}